\documentclass[conference]{IEEEtran}

\IEEEoverridecommandlockouts
\usepackage{cite}
\usepackage{amsmath,amssymb,amsfonts}
\usepackage{algorithmic}
\usepackage{graphicx}
\usepackage{textcomp}
\usepackage{xcolor}
\usepackage{float}
\usepackage{url}
\usepackage{subcaption}

\usepackage[colorlinks=true,linkcolor=black,citecolor=blue,urlcolor=blue]{hyperref}
\def\BibTeX{{\rm B\kern-.05em{\sc i\kern-.025em b}\kern-.08em
    T\kern-.1667em\lower.7ex\hbox{E}\kern-.125emX}}
\IEEEoverridecommandlockouts
\IEEEpubid{\begin{minipage}{\textwidth}\ \\[12pt]
979-8-3315-5937-3/25/\$31.00 \copyright2025 IEEE
\end{minipage}}
\begin{document}

\title{Deep Intrinsic Surprise-Regularized Control (DISRC): A Biologically Inspired Mechanism for Efficient Deep Q-Learning in Sparse Environments\\
}

\author{
Yash Kini$^{*}$ \\
James Madison High School \\
\texttt{kiniyash3@gmail.com}
\and
Shiv Davay$^{*}$ \\
Thomas Jefferson HSST\\
\texttt{davay.shiv@gmail.com}
\and
Shreya Polavarapu$^{*}$ \\
Northview High School \\
\texttt{shreyapolavarapu9@gmail.com}
}

\maketitle

\begin{abstract}
Deep reinforcement learning (DRL) has driven major advances in autonomous control. Still, standard Deep Q-Network (DQN) agents tend to rely on fixed learning rates and uniform update scaling, even as updates are modulated by temporal-difference (TD) error. This rigidity limits data efficiency and destabilizes convergence, especially in sparse-reward settings where feedback is infrequent. While methods like curiosity-driven exploration and prioritized replay indirectly influence learning dynamics, few approaches directly regulate update magnitudes using intrinsic signals. We introduce Deep Intrinsic Surprise-Regularized Control (DISRC), a biologically inspired augmentation to DQN that dynamically scales Q-updates based on latent-space surprise. DISRC encodes states via a LayerNorm-based encoder and computes a deviation-based surprise score relative to a moving latent setpoint. Each update is then scaled in proportion to both TD error and surprise intensity, promoting plasticity during early exploration and stability as familiarity increases. We evaluate DISRC on two sparse-reward MiniGrid environments, which included MiniGrid-DoorKey-8x8-v0 and MiniGrid-LavaCrossingS9N1-v0, under identical settings as a vanilla DQN baseline. In DoorKey, DISRC reached the first successful episode (reward $>$ 0.8) 33\% faster than the vanilla DQN baseline (79 vs. 118 episodes), with lower reward standard deviation (0.25 vs. 0.34) and higher reward area under the curve (AUC: 596.42 vs. 534.90). These metrics reflect faster, more consistent learning - critical for sparse, delayed reward settings. In LavaCrossing, DISRC achieved a higher final reward (0.95 vs. 0.93) and the highest AUC of all agents (957.04), though it converged more gradually. These preliminary results establish DISRC as a novel and effective mechanism for regulating learning intensity in off-policy agents, improving both efficiency and stability in sparse-reward domains. This positions DISRC as a step toward real-world decision systems, such as autonomous robots or clinical planners, where agents must adapt under sparse, delayed feedback. By treating surprise as an intrinsic learning signal, DISRC enables agents to modulate updates based on expectation violations, enhancing decision quality when conventional value-based methods fall short.
\end{abstract}


\begingroup
\renewcommand\thefootnote{*}
\footnotetext{These authors contributed equally to this work and share co-first authorship.}
\endgroup

\section{Introduction}
Deep Reinforcement Learning (DRL) has revolutionized sequential decision-making and control, empowering autonomous agents to operate in complex, high-dimensional domains such as robotic manipulation, autonomous navigation, and strategic gameplay~\cite{mnih2015human,pathak2017curiosity}. By leveraging deep neural networks to approximate value functions or policies, DRL overcomes the scalability bottlenecks of classical reinforcement learning~\cite{pathak2017curiosity}. Yet, despite impressive benchmarks, standard algorithms like Deep Q-Networks (DQNs) continue to suffer from critical limitations in sample efficiency, learning stability, and biological plausibility~\cite{pathak2017curiosity,schaul2015prioritized,gerstner2002spiking}.

Sample inefficiency stems from the vast number of interactions required to generalize effective behavior, while instability often results from bootstrapped targets and replayed experiences violating the independent and identically distributed (i.i.d.) assumption~\cite{schaul2015prioritized}. Though techniques such as curiosity-driven exploration~\cite{pathak2017curiosity} and Random Network Distillation (RND)~\cite{burda2018rnd} attempt to address sparse-reward regimes through intrinsic motivation, they primarily shape exploration rather than directly moderating how strongly the agent updates its policy in response to new experiences, instead supplementing the reward signal to foster exploration when external reinforcement is sparse. Similarly, Prioritized Experience Replay~\cite{schaul2015prioritized} enhances sample selection by emphasizing high temporal-difference (TD) errors but does not intrinsically modulate parameter update magnitudes. Adaptive optimizers like Adam and RMSprop~\cite{kingma2014adam,tieleman2012rmsprop} introduce gradient-based adjustment at the parameter level but lack a global agent-centered mechanism for regulating learning strength based on internal cognitive state.

In biological neural systems, we were motivated by adaptive plasticity where neuromodulators like dopamine scale learning rates based on perceived novelty and relevance. Therefore, we propose Deep Intrinsic Surprise-Regularized Control (DISRC), a novel augmentation to the DQN framework. DISRC introduces a biologically inspired self-regulatory mechanism that dynamically adjusts learning intensity through a latent-space surprise signal. Using a LayerNorm-based encoder, DISRC maps observations into an abstract latent space and maintains a moving setpoint that reflects the agent’s internal expectation. Surprise is computed as the normalized deviation from this setpoint and used to scale each Q-update in proportion to both TD error and surprise intensity. This enables the agent to remain highly plastic during early learning and exploration, while gradually adopting more conservative and stability-oriented updates as familiarity with the environment increases. In doing so, DISRC offers a new direction for stabilizing off-policy learning while aligning with neuroscientific principles of adaptive internal regulation~\cite{schultz1998predictive}.

We hypothesize that augmenting DQNs with an internal surprise signal to modulate update magnitude will improve learning efficiency and stability in sparse-reward environments. Specifically, we anticipate that DISRC agents will converge faster, exhibit lower reward standard deviation, and maintain higher cumulative reward relative to a vanilla DQN under identical training conditions.

\section{Methods}

\subsection{Environments}
To evaluate our proposed framework under sparse-reward conditions requiring compositional planning, we utilized two environments from the MiniGrid benchmark suite: \texttt{MiniGrid-DoorKey-8x8-v0} and \texttt{MiniGrid-LavaCrossingS9N1-v0}, both accessed via the \texttt{gymnasium-minigrid} interface. Each environment presents distinct challenges. In DoorKey, the agent must locate a key, unlock a door, and reach a goal, requiring temporal abstraction and memory of previous actions. In LavaCrossing, the agent must traverse a lava field with sparse safe paths to reach a distant goal, rewarding only precise spatial planning and long-term exploration. Observations consist of $7 \times 7 \times 3$ grid images representing the partial view of the environment centered on the agent.
All observations were flattened and normalized to $\in [0, 1]$ via a custom \texttt{preprocess\_obs()} function to enable compatibility with fully connected network architectures. Missions provided in text format were discarded for this study to isolate visual generalization.

\subsection{Baseline DQN Implementation}
As our baseline, we implemented a standard Deep Q-Network (DQN) agent using a multi-layer perceptron with two hidden layers of 256 units each, activated via ReLU and stabilized with Layer Normalization. The agent was trained using experience replay (buffer size: 50,000) and a soft target update mechanism ($\tau = 0.005$) to update a secondary target network.
The Q-values were optimized using the standard squared temporal-difference (TD) loss between predicted and bootstrapped target Q-values:
{\scriptsize
\begin{multline}
\mathcal{L}_{\text{DQN}} =
\mathbb{E}_{(s,a,r,s',d) \sim \mathcal{D}}
\big[ (Q(s,a) - (r + \gamma Q_{\text{target}}(s', \\
\arg\max_a Q(s',a))(1-d)))^2 \big]
\end{multline}
}
We used an Adam optimizer (learning rate $1\text{e}{-4}$), mini-batch size of 128, discount factor $\gamma = 0.99$, and $\epsilon$-greedy exploration with $\epsilon$ annealed from 1.0 to 0.1.
\subsection{DISRC Architecture}
Our proposed Deep Intrinsic Surprise-Regularized Control (DISRC) architecture augments the baseline DQN by introducing a biologically inspired mechanism that modulates learning based on internal surprise signals derived from abstracted latent states. DISRC introduces two major components: a LayerNorm-based encoder and a surprise-based controller.

\subsubsection{State Encoder}
Raw visual inputs are passed through a nonlinear encoder $f$, mapping the 147-dimensional observation vector to a 64-dimensional latent space. The encoder consists of:
\begin{itemize}
  \item Input Layer: Linear(147 $\rightarrow$ 256) + LayerNorm + ReLU
  \item Hidden Layer: Linear(256 $\rightarrow$ 128) + LayerNorm + ReLU
  \item Output Layer: Linear(128 $\rightarrow$ 64)
\end{itemize}

\subsubsection{Surprise Controller}
We compute an intrinsic surprise bonus using a deviation-based mechanism. A latent setpoint vector $\mu_t \in \mathbb{R}^{64}$ is maintained and updated over time as an exponential moving average of past encoded states. For each timestep, we compute the $L_2$ deviation between the normalized current latent state $s_t$ and $\mu_t$: 
$\text{deviation}_t = \big\lVert \tfrac{s_t}{\|s_t\|} - \tfrac{\mu_t}{\|\mu_t\|} \big\rVert_2$.

This deviation is scaled by a decaying coefficient $\beta_t = \beta_0 (1 -\text{progress}^{1.2})$, where $\text{progress} = t/T$ is the episode ratio. The resulting surprise penalty $b_t = -\beta_t \cdot \text{deviation}_t$ is added to the external reward after normalization:
$\hat{r}_t = \tfrac{r_t}{\text{EMA}(|r_t|)} + \lambda b_t$.
This reward modulation enables the agent to perform larger updates when encountering novel internal states, and progressively smaller updates as training stabilizes, supporting the core notion of surprise modification.

\begin{figure}[H]
  \centering
  \includegraphics[width=\linewidth]{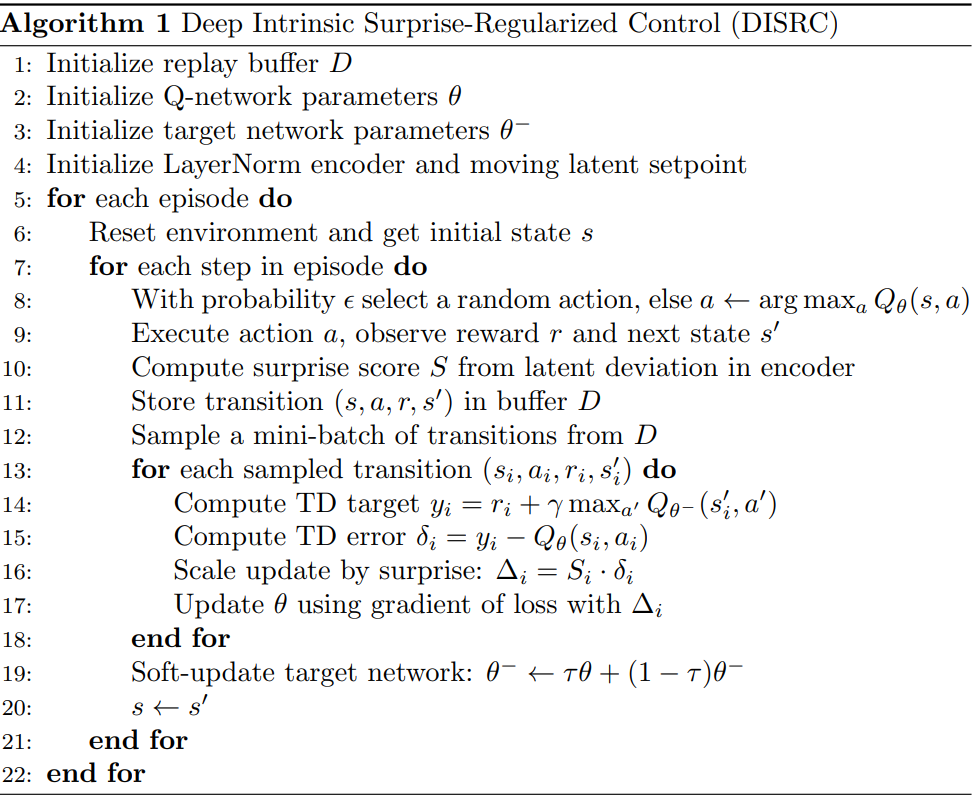}
  \caption{\textbf{DISRC Algorithm Pseudocode.} This diagram outlines the core loop of Deep Intrinsic Surprise-Regularized Control (DISRC). The agent dynamically adjusts Q-value update magnitudes using a biologically inspired surprise score computed from deviation relative to a moving latent setpoint.}
  \label{fig:disrc_algo}
\end{figure}
\subsection{Training Procedure}
Both the baseline and DISRC agents were trained independently on each environment using nearly precise training hyperparameters and experimental conditions, with the exception of the additional encoder, controller, and shaped reward logic in the DISRC model.

The experiments were conducted for 700 episodes in the DoorKey environment and 1200 episodes in the LavaCrossing environment. A replay buffer storing 50{,}000 transitions was used. The optimization was performed with Adam, using a learning rate of \(1 \times 10^{-4}\) for the Q-network and \(3 \times 10^{-4}\) for the encoder. Training employed a mini-batch size of 128, and the target network was updated softly with \(\tau = 0.005\). Exploration followed an \(\epsilon\)-greedy policy with linear decay, starting from \(\epsilon = 1.0\) and decaying to \(\epsilon_{\text{min}} = 0.1\). The loss function was the mean squared error of the TD error, and gradient norms were clipped between 0.2 and 0.3, depending on the environment.

\subsection{Evaluation Metrics}
We evaluated agent performance using several metrics collected across all episodes. The \textbf{Mean Final Reward} measures the average reward obtained over the last 50 episodes. The \textbf{Episodes to Threshold} metric records the number of episodes required for the agent to achieve a reward greater than 0.8. The \textbf{Loss Variance} quantifies the variance of the TD loss over all training steps, while the \textbf{Reward Standard Deviation} captures the global variability in episodic rewards. Finally, the \textbf{AUC (Area Under Reward Curve)} is computed as the trapezoidal integral over the episode reward sequence. Together, these metrics provide a comprehensive assessment of learning stability, efficiency, convergence speed, and behavioral variability.

\subsection{Code Availability}
The full implementation codebase is available at: \href{https://github.com/yashkini1/DISRC_URTC}{https://github.com/yashkini1/DISRC\_URTC}

\section{Results}

We evaluate the effectiveness of Deep Intrinsic Surprise-Regularized Control in comparison to a baseline Deep Q-Network (DQN) across two benchmark sparse-reward environments in MiniGrid: \texttt{MiniGrid-DoorKey-8x8-v0}, which requires sequential key-door navigation under partial observability, and \texttt{MiniGrid-LavaCrossingS9N1-v0}, which imposes high exploration risk due to misleading pathways and sparse terminal rewards.

\noindent\textbf{Table 1} presents a comparison of key metrics across both environments. DISRC demonstrates improved reward stability and learning efficiency, outperforming the vanilla DQN in mean reward, reward standard deviation, and area under the reward curve (AUC).

\begin{table*}[!t]
  \centering
  \includegraphics[width=0.8\textwidth]{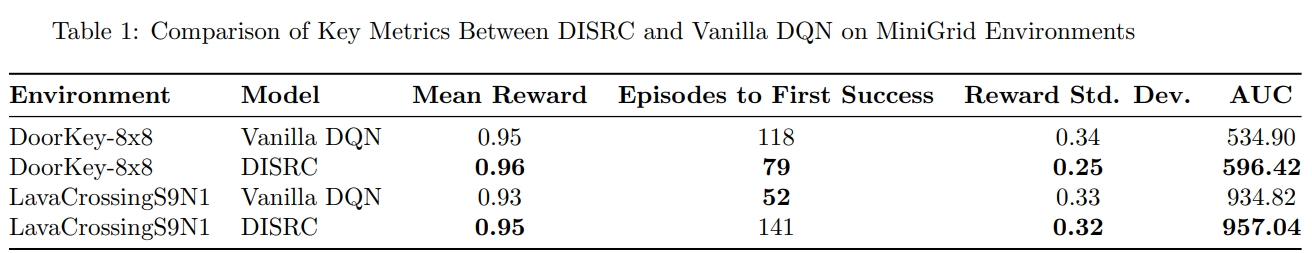}
  \label{tab:comparison}
\end{table*}

\begin{figure*}[!t]
  \centering
  \begin{subfigure}{0.49\linewidth}
    \centering
    \includegraphics[width=\linewidth]{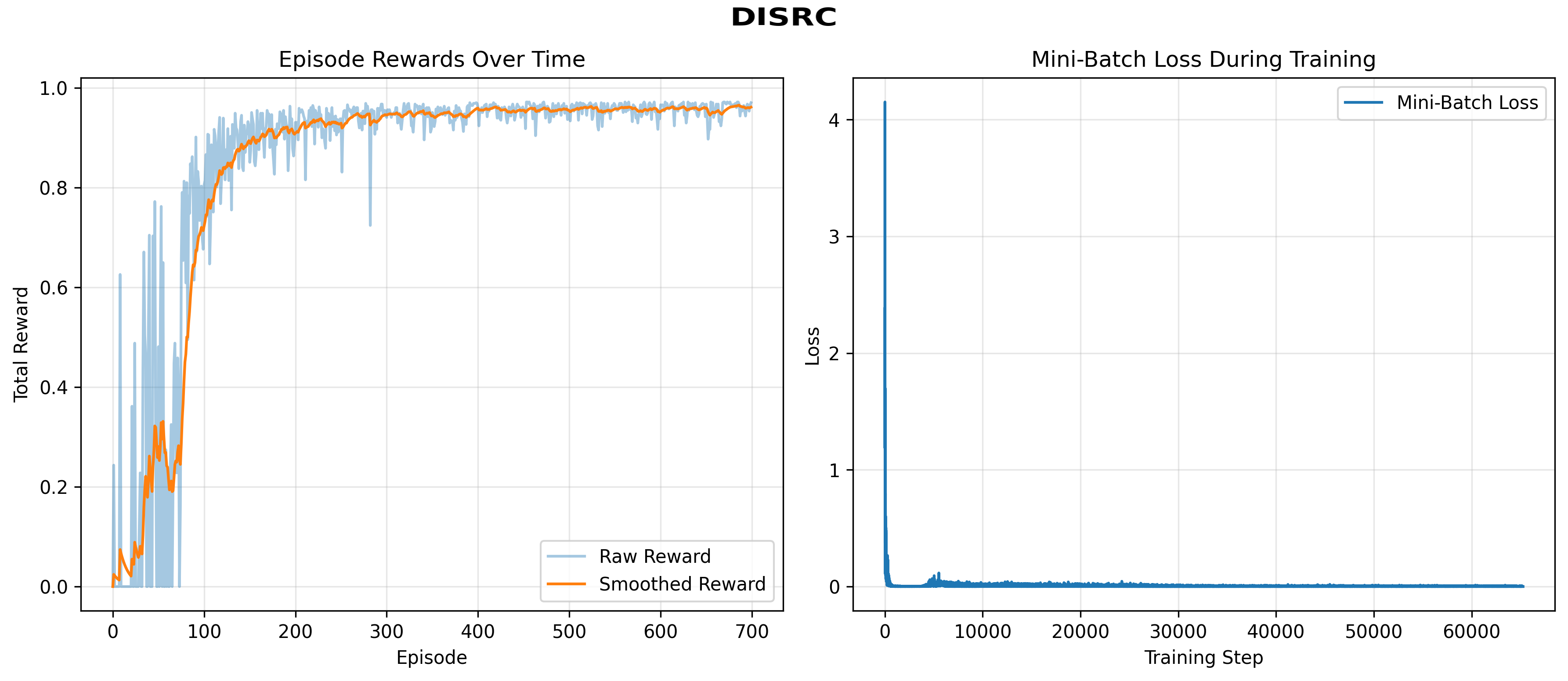}
    \caption{DISRC}
    \label{fig:doorkey_disrc}
  \end{subfigure}

  \begin{subfigure}{0.49\linewidth}
    \centering
    \includegraphics[width=\linewidth]{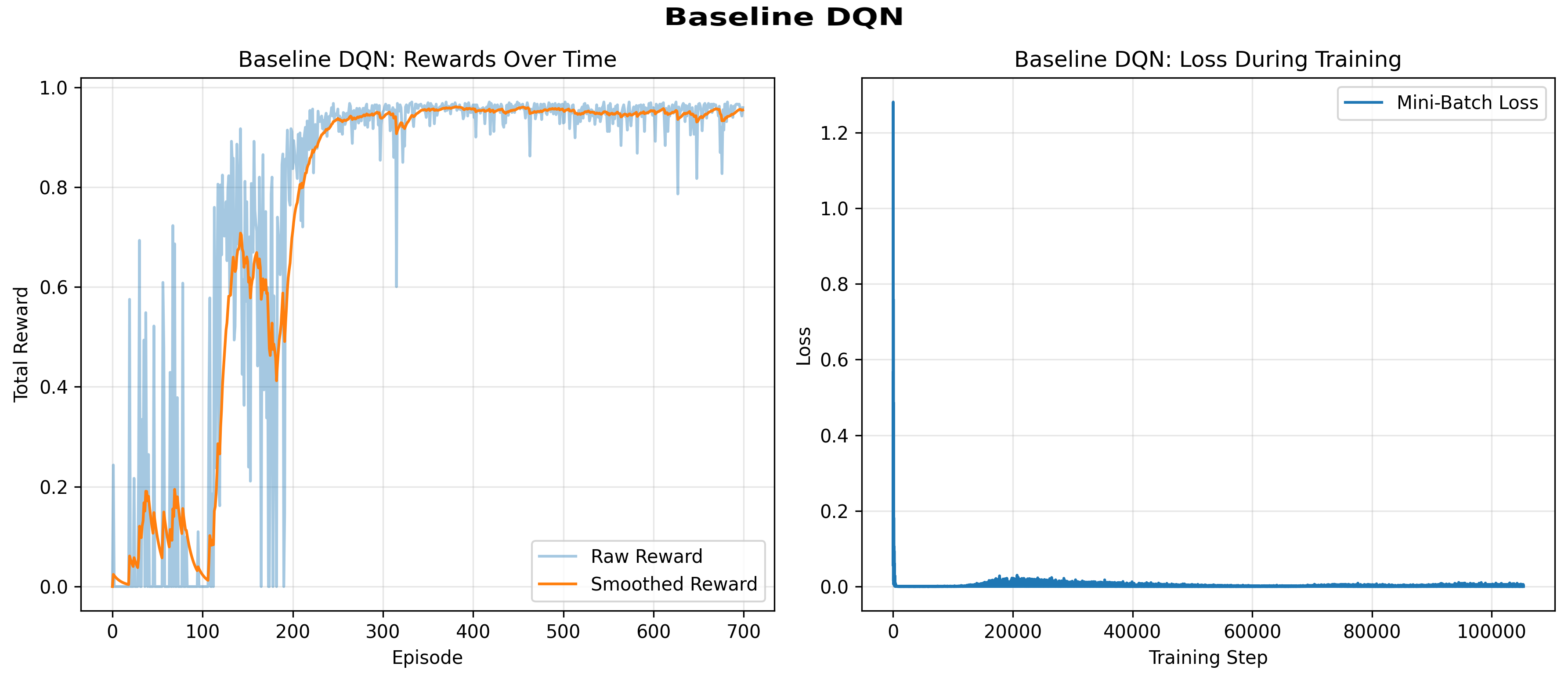}
    \caption{Baseline DQN}
    \label{fig:doorkey_dqn}
  \end{subfigure}

  \caption{\textbf{Learning Curves on DoorKey-8x8.} DISRC demonstrates more rapid convergence and smoother learning signals than baseline DQN, along with tighter loss distributions.}
  \label{fig:doorkey_curve}
\end{figure*}

\noindent\textbf{Figure~\ref{fig:doorkey_curve}} visualizes the learning curve for DoorKey-8x8. DISRC achieves faster convergence and smoother reward accumulation, maintaining reduced mini-batch loss compared to the baseline DQN.

In \texttt{DoorKey-8x8}, DISRC reached its first successful episode in only 79 episodes, a 33\% reduction compared to baseline DQN (118 episodes). Additionally, DISRC achieved a lower reward standard deviation (0.25 vs. 0.34) and a higher AUC (596.42 vs. 534.90), indicating both greater sample efficiency and increased behavioral stability. Though the final mean reward improvement is modest (0.96 vs. 0.95), the underlying learning trajectory (Figure~\ref{fig:doorkey_curve}) reveals accelerated convergence, sustained reward acquisition, and a consistently smoother reward signal. The corresponding loss curves further demonstrate that DISRC maintains tighter mini-batch loss distributions and faster adjustments of early instability, suggesting improved representational conditioning and reduced variance in Q-updates.

\begin{figure*}[!t]
  \centering
  \begin{subfigure}{0.49\linewidth}
    \centering
    \includegraphics[width=\linewidth]{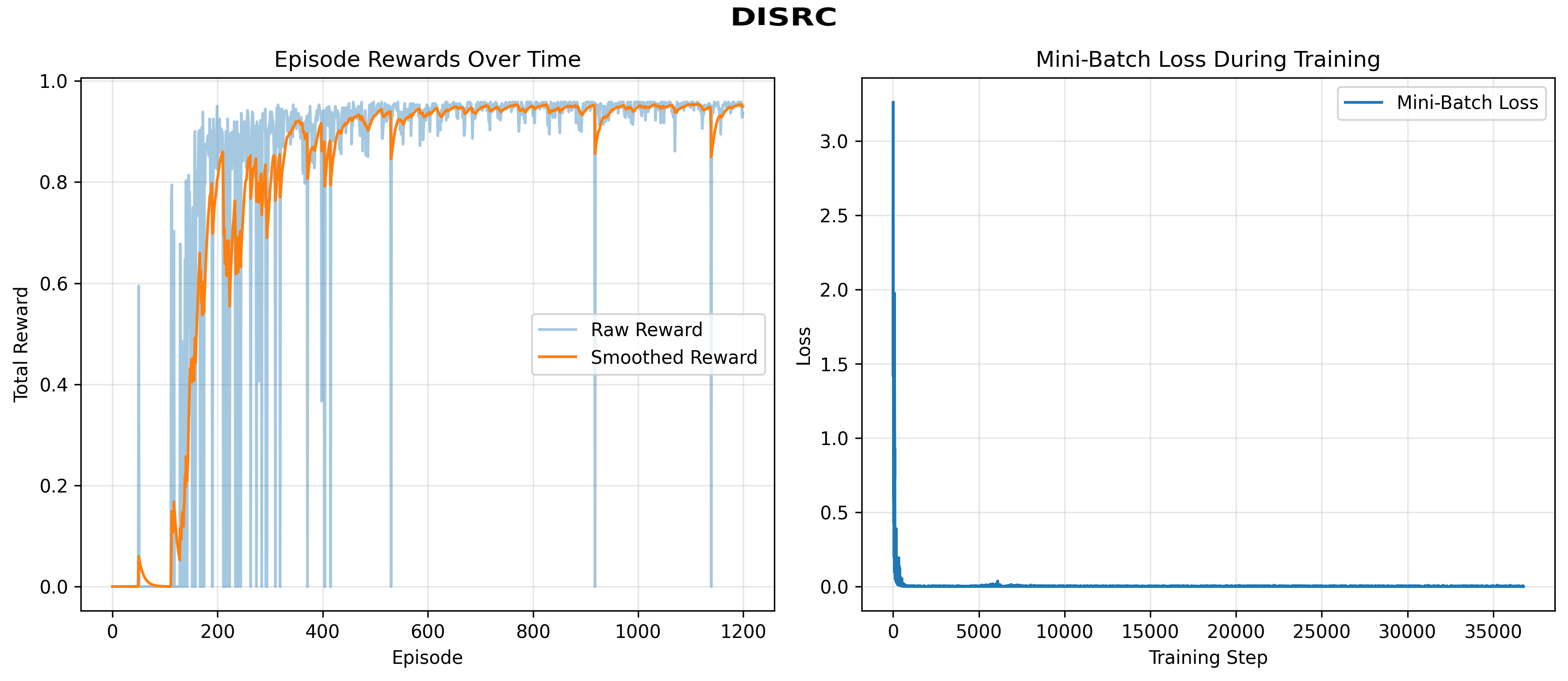}
    \caption{DISRC}
    \label{fig:lava_disrc}
  \end{subfigure}

  \begin{subfigure}{0.49\linewidth}
    \centering
    \includegraphics[width=\linewidth]{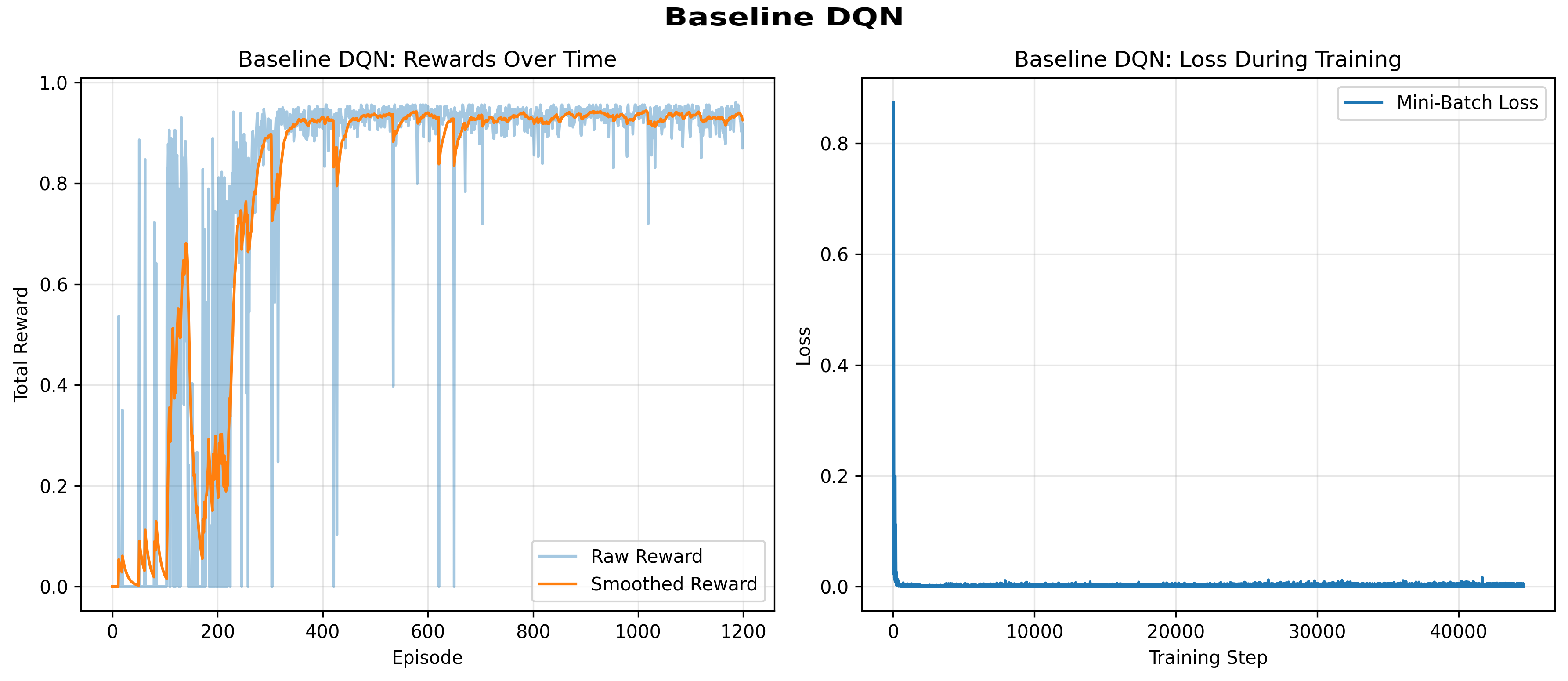}
    \caption{Baseline DQN}
    \label{fig:lava_dqn}
  \end{subfigure}

  \caption{\textbf{Learning Curves on LavaCrossingS9N1.} DISRC exhibits greater reward consistency and training stability under sparse, high-risk transitions.}
  \label{fig:lava_curve}
\end{figure*}

\noindent\textbf{Figure~\ref{fig:lava_curve}} presents learning curves for \texttt{LavaCrossingS9N1}. In the sparse-reward LavaCrossingS9N1 environment, DISRC sustains competitive performance with lower reward volatility and more stable training loss, despite requiring more episodes to reach the first success.

In \texttt{LavaCrossingS9N1}, an environment characterized by high-risk dead-ends and deceptive paths, DISRC continued to demonstrate strong generalization. It achieved a higher final mean reward (0.95 vs. 0.93) and greater overall efficiency, as evidenced by an AUC of 957.04 compared to the baseline's 934.82. While the number of episodes to first success was longer (141 vs. 52), this delay reflects a more consistent learning pattern with lower reward standard deviation (0.32 vs. 0.33), avoiding the reward collapse and overcommitment to suboptimal policies observed in the baseline. These dynamics are portrayed in Figure~\ref{fig:lava_curve}, which again shows that DISRC exhibits more consistent progression and tighter loss trajectories, even in the presence of sparse, noisy feedback.

To clarify the mechanics driving these gains, Figure~\ref{fig:disrc_algo} provides a high-level pseudocode overview of the DISRC training pipeline. Unlike conventional DQN variants that regulate updates solely via TD error or heuristics like prioritized replay~\cite{schaul2015prioritized}, DISRC introduces a deviation-based surprise modulation term derived from the encoder's latent representation. This biologically inspired mechanism dynamically adjusts Q-update magnitudes based on distance from a moving internal setpoint, enabling a phase transition from high plasticity in early exploration to conservative updates during policy consolidation.

Together, these findings validate the central hypothesis of this work: that DISRC adds a previously unexplored axis of internal regulation to off-policy reinforcement learning agents. By converting intrinsic surprise into an adaptive scaling factor for updates, DISRC not only improves stability and efficiency in sparse-reward domains, but also models a core cognitive mechanism - the reweighting of experience by deviation from the expectation - that is prevalent in biological systems. This positions DISRC as a foundational and promising step toward building reinforcement learners that learn not just from what they do, but from how unexpected it felt to do it.

\section{Discussion}

The empirical performance of DISRC across both DoorKey and LavaCrossing environments shines light on a compelling observation: when extrinsic rewards are sparse or deceptive, the internal modulation of learning plasticity via latent surprise yields measurable gains in sample efficiency, stability, and convergence quality. In contrast to environments with dense and consistent reward signals such as CartPole - where traditional Q-learning or adaptive optimizers may suffice - sparse domains demand a more introspective learning dynamic. By conditioning the update magnitude on the deviation from a moving latent setpoint, DISRC functions as an intrinsic calibration mechanism, enhancing the agent’s sensitivity to rare but meaningful transitions and reducing premature convergence to suboptimal policies. This aligns with neurophysiological evidence on dopamine-based error signaling in the brain, where unexpected outcomes modulate synaptic updates more intensely than routine events.

While prior architectures such as curiosity-driven exploration~\cite{pathak2017curiosity}, random network distillation~\cite{burda2018rnd}, and prioritized experience replay~\cite{schaul2015prioritized} attempt to steer exploration or replay frequency, DISRC introduces a fundamentally orthogonal mechanism - what we term \textit{second-order update regulation}. Rather than shaping which experiences are seen or how rewards are interpreted, DISRC directly scales how much the agent learns from a given experience. This self-regulatory dimension is loosely comparable to neuromodulatory systems in biological agents, which adjust learning rates based on contextual novelty and internal uncertainty. As a result, DISRC draws on principles from cognitive neuroscience, particularly dopaminergic surprise signaling, offering a novel framework for informing a practical reinforcement learning strategy.

However, this design is not without its trade-offs and limitations. The reliance on latent setpoint tracking introduces additional computation cost during forward passes and requires careful tuning of hyperparameters such as the learning rate for the latent moving average and the decay exponent on episode ratio. Moreover, as with any architecture embedding internal state dynamics, the relationship between encoder representation fidelity and surprise estimation must be finely balanced to avoid signal drift or noisy modulations. Additionally, while these findings support our hypothesis, we were surprised by DISRC's performance in LavaCrossing, where slower initial success was offset by stronger long-term generalization - an outcome that merits further analysis over a wider range of environments and baselines. Subsequently, our evaluation was limited to MiniGrid environments to ensure experimental clarity and consistency; we selected these tasks for their well-defined structure, reproducibility, and alignment with the objectives of surprise-modulated learning. These limitations signal clear next steps: extending DISRC to partially observable environments with high-dimensional visual or proprioceptive inputs, benchmarking against explainability-augmented DRL baselines, and evaluating hybrid frameworks that combine intrinsic update modulation with curiosity-driven intrinsic rewards for deeper policy shaping.

The implications of this study reach beyond simulated benchmarks. In real-world decision systems - such as autonomous robotics, clinical treatment planning, and adaptive network control - agents often operate with incomplete feedback and delayed consequence structures. DISRC’s biologically grounded update modulation offers a pathway toward systems that can not only learn from delayed reward, but also recognize when to update cautiously, aggressively, or not at all, a feature that is critically missing from most value-based methods. By formalizing surprise as a first-class signal for learning regulation, this work lays the foundation for next-generation DRL agents that reason not only over outcomes but also over their expectations.

In summary, DISRC represents a novel and generalizable mechanism for internal update regulation in reinforcement learning. It reframes surprise not as an exploration driver or curiosity, but as a principled signal for adaptive learning control. Through this lens, we move one step closer to agents that mirror biological adaptability - not just in action but in cognition. These results open promising avenues for both scientific inquiry and real-world deployment in autonomous, high-stakes environments.

\section*{Acknowledgments}

This work was conducted through the George Mason University Aspiring Scientists Summer Internship 2025 Program (ASSIP). We thank the Farama Foundation for providing the MiniGrid benchmark suite and the IEEE URTC organizing committee for the opportunity to submit and present.

\vspace{5pt}
\color{red}
\textit{All figures and empirical data were generated by the authors.}

\begin{thebibliography}{99}

\bibitem{mnih2015human}
V.~Mnih, K.~Kavukcuoglu, D.~Silver, et al., “Human-level control through deep reinforcement learning,” \emph{Nature}, vol. 518, no. 7540, pp. 529–533, 2015. \url{https://www.nature.com/articles/nature14236}

\bibitem{pathak2017curiosity}
D.~Pathak, P.~Agrawal, A.~A.~Efros, and T.~Darrell, “Curiosity-driven exploration by self-supervised prediction,” arXiv:1705.05363, 2017. \url{https://arxiv.org/abs/1705.05363}

\bibitem{burda2018rnd}
Y.~Burda, H.~Edwards, A.~Storkey, and O.~Klimov, “Exploration by random network distillation,” arXiv:1810.12894, 2018. \url{https://arxiv.org/abs/1810.12894}

\bibitem{schaul2015prioritized}
T.~Schaul, J.~Quan, I.~Antonoglou, and D.~Silver, “Prioritized experience replay,” arXiv:1511.05952, 2015. \url{https://arxiv.org/abs/1511.05952}

\bibitem{kingma2014adam}
D.~P.~Kingma and J.~Ba, “Adam: A method for stochastic optimization,” arXiv:1412.6980, 2014. \url{https://arxiv.org/abs/1412.6980}

\bibitem{tieleman2012rmsprop}
T.~Tieleman and G.~Hinton, “RMSprop: Divide the gradient by a running average of its recent magnitude,” Coursera: Neural Networks for Machine Learning, 2012. \url{https://www.cs.toronto.edu/~tijmen/csc321/slides/lecture_slides_lec6.pdf}

\bibitem{gerstner2002spiking}
W.~Gerstner and W.~M.~Kistler, \emph{Spiking Neuron Models: Single Neurons, Populations, Plasticity}. Cambridge University Press, 2002. \url{https://doi.org/10.1017/CBO9780511815706}

\bibitem{schultz1998predictive}
W.~Schultz, “Predictive reward signal of dopamine neurons,” \emph{Journal of Neurophysiology}, vol. 80, no. 1, pp. 1–27, 1998. \url{https://pubmed.ncbi.nlm.nih.gov/9658025/}

\bibitem{cheng2025survey}
Z.~Cheng, J.~Yu, and X.~Xing, “A survey on explainable deep reinforcement learning,” arXiv:2502.06869, 2025. \url{https://arxiv.org/abs/2502.06869}

\bibitem{liang2024episodic}
D.~Liang, Y.~Zhang, and Y.~Liu, “Episodic reinforcement learning with expanded state-reward space,” \emph{Proc. of the 23rd International Conference on Autonomous Agents and Multiagent Systems (AAMAS)}, 2024. \url{https://arxiv.org/abs/2401.10516}

\end{thebibliography}
\end{document}